\documentclass[sigconf,natbib,anonymous=false]{acmart}
\pdfoutput=1
\acmSubmissionID{660}
\usepackage{acronym}
\usepackage[skip=0pt]{caption}
\usepackage[inline]{enumitem}
\usepackage{booktabs}
\usepackage{makecell}
\usepackage{multirow}
\usepackage{subcaption}
\usepackage{xspace}
\usepackage{adjustbox}
\usepackage{bbding}
\usepackage{pifont}
\usepackage{caption}
\usepackage{graphicx}
\usepackage{float} 
\usepackage{subcaption}
\usepackage{xspace}
\usepackage{enumitem}[inline]
\usepackage[ruled,linesnumbered]{algorithm2e}
\usepackage{hyperref}

\AtBeginDocument{%
  \providecommand\BibTeX{{%
    \normalfont B\kern-0.5em{\scshape i\kern-0.25em b}\kern-0.8em\TeX}}}

\newcommand{\header}[1]{\vspace*{2mm}\noindent\textbf{#1}}

\acrodef{IR}{information retrieval}
\acrodef{AMT}{Amazon Mechanical Turk}
\acrodef{HIT}{human intelligence task}
\acrodef{CQA}{conversational question-answering}
\acrodef{QuAC}{Question Answering in Context}
\acrodef{LLM}{large language model}
\acrodef{RC}{reading comprehension}

\newcommand{\teacherSim}{{\textit{teacher\textsubscript{Sim}}}\xspace}
\newcommand{\studentSim}{{\textit{student\textsubscript{Sim}}}\xspace}
\newcommand{\pageTitle}{{$t$}\xspace}
\newcommand{\background}{{$b$}\xspace}

\newcommand{\ourdataset}{SimQuAC\xspace}
\newcommand{\sectionHeader}{{$h$}\xspace}
\newcommand{\textSection}{{$s$}\xspace}

\newcommand{\answerQUAC}{{\textit{answer\textsubscript{QuAC}}}\xspace}
\newcommand{\answerSim}{{\textit{answer\textsubscript{Sim}}}\xspace}
\newcommand{\quac}{{QuAC}\xspace}

\newcommand{\QGfunc}{{$\phi_S$}\xspace}
\newcommand{\AGfunc}{{$\phi_T$}\xspace}

\newcommand{\ValidTeacher}{{$\sigma_T$}\xspace}
\newcommand{\ValidStudent}{{$\sigma_S$}\xspace}

\newcommand{\promptSelectionTeacher}{{$\omega_T$}\xspace}
\newcommand{\promptSelectionStudent}{{$\omega_S$}\xspace}
\newcommand{\teacherPrompt}{{\textit{Instruction\textsubscript{T}}}\xspace}
\newcommand{\studentPrompt}{{\textit{Instruction\textsubscript{S}}}\xspace}

\newcommand{\teacher}{\texttt{teacher}\xspace}
\newcommand{\student}{\texttt{student}\xspace}

\newcommand{\students}{\texttt{students}\xspace}

\copyrightyear{2024}
\acmYear{2024}
\setcopyright{rightsretained}
\acmConference[WSDM '24]{Proceedings of the 17th ACM International Conference on Web Search and Data Mining}{March 4th-8th, 2024}{Mérida, Yucatán}
\acmBooktitle{Proceedings of the 17th ACM International Conference on Web Search and Data Mining (WSDM '24), March 4th-8th, 2024, Mérida, Yucatán}
\acmDOI{}
\acmISBN{}

\author{Zahra Abbasiantaeb}
\affiliation{%
  \institution{University of Amsterdam}
  \city{Amsterdam}
  \country{The Netherlands}
}
\email{z.abbasiantaeb@uva.nl}

\author{Yifei Yuan}
\affiliation{%
  \institution{The Chinese University of Hong Kong}
  \city{Hong Kong}
  \country{Hong Kong SAR}}
\email{yfyuan@se.cuhk.edu.hk}

\author{Evangelos Kanoulas}
\affiliation{%
  \institution{University of Amsterdam}
  \city{Amsterdam}
  \country{The Netherlands}
}
\email{e.kanoulas@uva.nl}

\author{Mohammad Aliannejadi}
\affiliation{%
  \institution{University of Amsterdam}
  \city{Amsterdam}
  \country{The Netherlands}
}
\email{m.aliannejadi@uva.nl}
\usepackage{subcaption}
\usepackage{xspace}
\usepackage{graphicx}
\usepackage{caption}
\usepackage{subcaption}
\usepackage{multirow}
\usepackage{xcolor}

\definecolor{color1}{rgb}{0.2, 0.81, 0.30}
\definecolor{color2}{rgb}{0.66, 0.25,0.97}
\definecolor{color3}{rgb}{0.92, 0.4, 0.08}
\definecolor{color4}{rgb}{0.44, 0.4, 0.9}

\allowdisplaybreaks

\begin{document}

\title[]{Let the LLMs Talk: Simulating Human-to-Human Conversational QA via Zero-Shot LLM-to-LLM Interactions}

\begin{abstract}
\Ac{CQA} systems aim to create interactive search systems that effectively retrieve information by interacting with users. To replicate human-to-human conversations, existing work uses human annotators to play the roles of the questioner (\student) and the answerer (\teacher). Despite its effectiveness, challenges exist as human annotation is time-consuming, inconsistent, and not scalable. To address this issue and investigate the applicability of \acp{LLM} in \ac{CQA} simulation, we propose a simulation framework that employs zero-shot learner \acp{LLM} for simulating \teacher--\student interactions. Our framework involves two LLMs interacting on a specific topic, with the first LLM acting as a \student, generating questions to explore a given search topic. The second LLM plays the role of a \teacher by answering questions and is equipped with additional information, including a text on the given topic. We implement both the \student and \teacher by zero-shot prompting the GPT-4 model.
To assess the effectiveness of LLMs in simulating CQA interactions and  understand the disparities between LLM- and human-generated conversations, we evaluate the simulated data from various perspectives. We begin by evaluating the \teacher's performance through both automatic and human assessment. Next, we evaluate the performance of the \student, analyzing and comparing the disparities between questions generated by the LLM and those generated by humans. Furthermore, we conduct extensive analyses to thoroughly examine the LLM performance by benchmarking state-of-the-art reading comprehension models on both datasets. Our results reveal that the \teacher LLM generates lengthier answers that tend to be more accurate and complete. The \student LLM generates more diverse questions, covering more aspects of a given topic.
\end{abstract}


\maketitle

\acresetall

\vspace{-5mm}
\section{Introduction}
Over the years, the information retrieval (IR) community has strived to create an interactive and iterative search system that effectively retrieves information~\cite{Croft1987I3RAN,Kotov2010TowardsNQ,Belkin1995CasesSA,aliannejadi2019asking}. Recent advancements in \acf{CQA} systems have been successful in achieving this goal by retrieving relevant information and engaging in back-and-forth interactions with users to fully understand their information needs~\cite{Qu2020OpenRetrievalCQ,Reddy2018CoQAAC}. Under this case, existing work captures the iterative dynamics of conversations, where a set of annotators play the role of the questioner (\student) and the answerer (\teacher) over a pre-defined search topic~\cite{Li2016LearningTD,Choi2018QuACQA,Trischler2016NewsQAAM}. 

Despite the effectiveness of previous efforts in this task, several drawbacks exist. One major challenge is the maintenance of a large team of annotators to generate a substantial number of conversations. This process can be time-consuming, resource-intensive, and expensive. Additionally, relying solely on human annotators may introduce variations in the quality and consistency of the generated conversations. 
Also, in many cases, the human \student cannot effectively explore a given topic that is out of their background knowledge. For example, a person who has expertise in geography can better explore a related topic rather than a person who does not. In contrast,  \acfp{LLM} can leverage their vast background knowledge to effectively play the role of a geography expert in a conversation. Therefore, it is crucial to explore automated approaches that can generate simulated conversations, reducing the dependency on human annotators and making the process more efficient and scalable. 

User simulation is an important emerging research frontier for conversational search development and evaluation~\cite{balog2021conversational,owoicho2023exploiting}, where the focus mainly is on simulating the user behavior under a certain condition, such as responding to system's actions~\cite{terragni2023incontext}, answering clarifying questions~\cite{sekulic2022evaluating}, and giving feedback on system answer~\cite{owoicho2023exploiting}. The main drawback of existing research on user simulation is its reactive nature, where the simulated user just passively respond to the system's utterance. In real-world scenarios, however, users' actions are a mix of proactive and reactive actions, initiating and frequently guiding conversations by posing questions that stem from their underlying information need.

In this work, we aim to explore \acp{LLM}' effectiveness in simulating a proactive user, exploring a pre-defined topic in a conversational setting. To this aim, we replicate the \teacher--\student conversational simulation adopted by \citet{Choi2018QuACQA} while replacing both human parties with \acp{LLM}, enabling us to effectively evaluate and compare the performance of \acp{LLM} with human annotators. This leads us to our first research question, \textbf{RQ1:} \textit{how can we employ \acp{LLM} to generate such simulated conversations effectively and automatically?} 
We answer this question by proposing a zero-shot \ac{LLM}-to-\ac{LLM} simulation framework where the \student \ac{LLM} aims to explore a topic by posing various questions and the \teacher \ac{LLM}'s goal is to provide complete and correct answers to the questions. We implement both the \student and \teacher by zero-shot prompting GPT-4. \looseness=-1

The usage of \acp{LLM} in this setting leads us to the next research questions \textbf{RQ2:} \textit{how can we evaluate the role of \acp{LLM} in CQA simulation?} and \textbf{RQ3:} \textit{how do \ac{LLM}- and human-generated conversations compare?} To address these questions: 
\begin{enumerate}[nosep, leftmargin=*, label=(\roman*)]
    \item We first conduct an extensive independent evaluation of the \teacher, measuring its effectiveness in this task. To this aim, we conduct an extensive human evaluation task where the annotators compare \ac{LLM}- and human-generated answers on the same questions side by side. 
    \item We then evaluate the performance of the \student. To this aim, we compare the patterns and question-asking behavior of the \ac{LLM} and human from various perspectives, discovering interesting patterns. For example, we find that \ac{LLM}-generated questions lead to more topical coverage.
    \item Finally, we conduct extensive analyses to thoroughly examine the performance of the LLM by benchmarking state-of-the-art reading comprehension models on both datasets.
\end{enumerate}
We find that \ac{LLM}-generated answers are generally lengthier and more comprehensive. Also, they are more consistent and fluent. Moreover, our human evaluation reveals that the \ac{LLM} \teacher is more accurate in providing correct answers. Upon benchmarking state-of-the-art reading comprehension models, we find that pre-trained models exhibit more effective performance on \ac{LLM}-generated data. This efficacy may result from certain biases in the generated conversations and the enhanced consistency within it. 

Overall, our contributions can be summarized as follows:
\begin{itemize}[nosep, leftmargin=*]
    \item We leverage \acp{LLM} to mimic human-to-human interaction in a CQA setting using zero-shot prompting. We prompt two \acp{LLM} to conduct \teacher--\student simulation and propose an \ac{LLM}-generated dataset, called \ourdataset.\footnote{Code and data available at \url{https://github.com/ZahraAbbasiantaeb/SimQUAC.git}}
    \item We propose and perform a comprehensive automatic and human evaluation framework, as well as linguistic analysis to evaluate the \ac{LLM}'s effectiveness in this setting on the \teacher and \student level.
    \item We conduct extensive analyses on \ac{LLM}- and human-generated conversations, discovering many interesting patterns exhibited by humans and \acp{LLM} during \ac{CQA}.
\end{itemize}

\begin{figure}[t]
    \centering
    \includegraphics[width=0.45\textwidth]{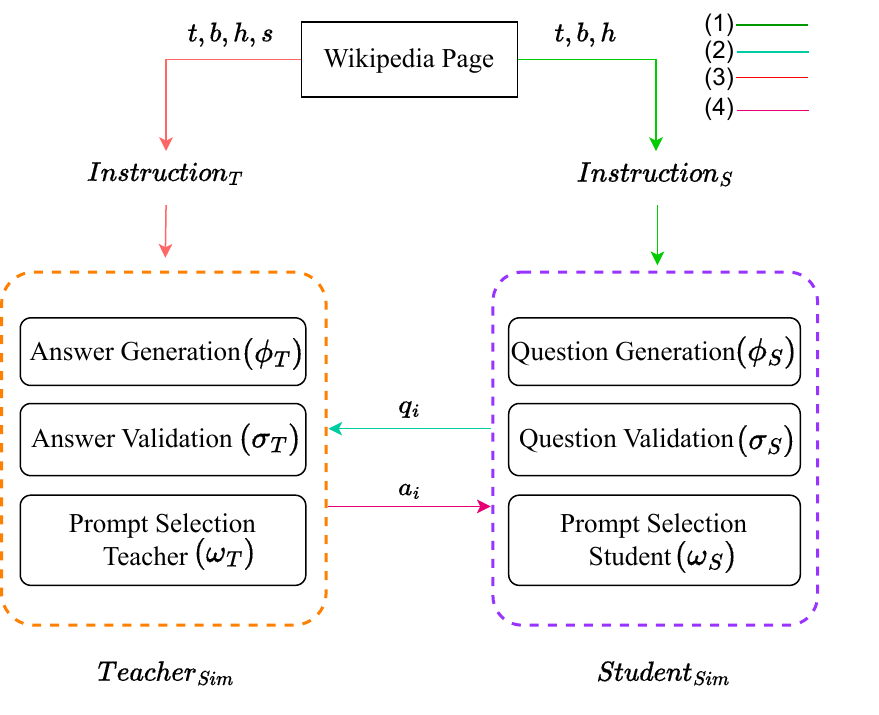}
    \caption{A high-level view of the architecture of our Simulation framework.}
    \label{fig:flowchart}
\end{figure}

\section{Methodology}
\label{method}
\subsection{Problem setting}
\label{quacintro}
Our experimental setup involves simulating an information-seeking conversation, where a \student interacts with a \teacher  in a question-answering conversation.  
Based on that, we adopt the setting established by the \acf{QuAc} dataset~\citep{Choi2018QuACQA}, which serves as a widely recognized benchmark for evaluating the effectiveness of CQA models. The dataset revolves around discussions based on Wikipedia articles. It consists of conversation contexts where a crowdworker plays the role of a questioner (\student) and engages in a conversation with another crowdworker who acts as an answerer (\teacher). Specifically, the \teacher is given access to the entire Wikipedia section article and aims to generate responses to questions posed by the \student. To ensure fairness, the \teacher 's responses are limited to selecting the appropriate answer span from the article. In contrast, the \student is only provided with the article's title and tries to use this limited information to ask relevant questions and explore the topic. As the \students engage in the conversation, they explore the topic by asking questions, and the conversation unfolds accordingly.

\subsection{Task formulation}
\label{taskform}
As is mentioned in Section \ref{quacintro},  the conversation evolves around a Wikipedia article titled \pageTitle. The \student is only provided with limited access to information, including the section header $h$ as information need and the first paragraph of the main article \background, which serves as the background information. The \teacher, on the other hand, has access to the additional information, including the full text of the section \textSection. The conversation begins when the \student raises an initial question $q_0$ and the teacher provides an answer, denoted as $a_0$. After receiving the answer from \teacher, \student continues to ask more questions until some stoppage criteria are met. Specifically, following previous work~\cite{Choi2018QuACQA,Trischler2016NewsQAAM,Rajpurkar2016SQuAD1Q}, instead of answering with free-text, the \teacher must select one or several contiguous spans from text as the answer. 
Note that although limiting the LLM to select text spans restricts the \teacher's ability to freely provide answers, it offers the advantage of simplified answer evaluation and prevents hallucination. This setting enables us to examine the proficiency of \acp{LLM} in tasks like CQA and \ac{RC} by comparing their performance against existing methods.

\subsection{Model framework overview}
In order to have a better understanding of \textbf{RQ1}, we propose a LLM-based framework. Figure~\ref{fig:flowchart} illustrates the overall architecture of our model, showcasing the interactions between the two \acp{LLM}. The entire process evolves around a Wikipedia page. The purple box on the right plots the simulated \student, named \studentSim{}, while the orange box on the left plots the simulated \teacher, named \teacherSim{}. 
The \teacherSim and \studentSim contain several components to generate acceptable answers and questions, respectively.

The process of generating the conversation starts with initializing \studentSim by giving the instruction prompt \studentPrompt. 
The \studentPrompt prompt aims to guide the \student \ac{LLM} \studentSim in generating the first question $q_0$ in the \textit{question generation} component (\QGfunc). Then, we pass the generated question $q_0$ to the \textit{question validation} component (\ValidStudent). This component plays a critical role in ensuring the structural integrity of the generated questions. If it determines that the structure of the question is not acceptable, \studentSim will prompt \QGfunc again to regenerate $q_0$.
After that, we forward $q_0$ to \teacherSim, which concatenates it with the instruction prompt \teacherPrompt, forming a combined input. This combined input is then fed to the \textit{answer generation} component (\AGfunc) for generating the answer $a_0$. To ensure that the generated answers adhere to our defined setting (i.e., corresponding to one or multiple segments in the section text \textSection), an \textit{answer validation} (\ValidTeacher) component is leveraged to check the validity of $a_0$. 
If $a_0$ is determined as an invalid answer, a \textit{prompt selection teacher} component \promptSelectionTeacher will select the appropriate prompt ($p_{T}$) and pass it to \AGfunc to regenerate the answer $a_0$. 
This step continues when $a_0$ is determined to be valid by \ValidTeacher, it is passed back to \studentSim{}. 

Similarly, \studentSim{} incorporates the \textit{prompt selection for student} component (\promptSelectionStudent{}) to select the optimal prompt $p_S$ for generating the subsequent question $q_i$. Once chosen, it transfers $p_S$ to the question generation \QGfunc{} module again, where $p_S$ is then employed to generate the next question $q_i$.
This back-and-forth question--answering process continues until the stoppage criteria are met. In each turn, the generated question $q_i$ and answer $a_i$ will be stored. Algorithm \ref{alg:simulation} shows the detailed simulation process of our model. In the following sections, we will provide detailed explanations of each component to further elucidate its functionality.

\RestyleAlgo{ruled}
\SetKwComment{Comment}{/* }{ */}

\begin{algorithm}
\caption{Data simulation algorithm}\label{alg:simulation}

\KwData{\pageTitle, \background, \textSection, \sectionHeader, \ $N$, $patience$}
\KwResult{$q_{0 \dots N}$, $a_{0 \dots N}$}
\studentPrompt{}$\gets $ StudentInitialPrompt(\pageTitle, \background, \sectionHeader)\;
\teacherPrompt{} $\gets $ TeacherInitialPrompt(\pageTitle, \background, \sectionHeader, \textSection, $q_0$)\;
$i \gets 0$ \;
\While{$i < N$}{
    $m \gets$ patience \;
    \eIf{$i==0$}{
        {$q_i \gets$ \QGfunc(\studentPrompt{}) \Comment*[r]{\studentSim}}}
   { 
        {$p_S \gets $ \promptSelectionStudent($a_i$)}\;
       {$q_i \gets $ \QGfunc($p_S$)}\;
       }

  \While{\ValidStudent($q_i$) is False}{
  $p_S \gets$ updatePromptToAskShortQuestion($p_S$)\;
    $q_i \gets $ \QGfunc($p_S$)\;
  }
  
    \eIf{$i==0$} {
    $a_i \gets$  \AGfunc(\teacherPrompt) \Comment*[r]{\teacherSim}}
    {$a_i \gets$ \AGfunc($q_i$)}
    
    \While{\ValidTeacher($a_i$, \background, \textSection) is False and $m>0$}{ 
    $p_T \gets $ \promptSelectionTeacher($a_i$, \background, \textSection)\;
    $a_i \gets$ \AGfunc($p_T$)\; 
    decrement m\;
  }
  increment $i$\;
  
}
\end{algorithm}

\begin{table*}[]
    \centering
        \caption{The template for constructing \teacherPrompt (left side) and the \studentPrompt (right side). The variables inside ``[ ]'' would be filled based on the input Wikipedia page. }
    \begin{tabular}{|p{6.7cm}|p{10.3cm}|}
    \hline
     \centering  \teacherPrompt & \centering\arraybackslash   \studentPrompt
     \\ \hline
        & \\
        Topic: [$t$] \newline
     Background knowledge [$b$] \newline \newline
        \textcolor{color3}{In this task, you will be given a text about the topic explained above. You will answer my questions from this text.  Please remember that you cannot generate the answer on your own but should only copy a continuous span from the original text and the copied answer should not exceed 40 tokens.  If you cannot find the answer in the text, please generate ‘I cannot find the answer’.} \newline \newline
        Section header: [$h$]  \newline
        Section text: [$s$]
       &  
\textcolor{color4}{In this task, I am a teacher and have a document, you are a curious \student who wants to explore this document by asking questions.
The main objective is to learn most of the documents that I have. I will explain to you the topic and background knowledge of the document. Then I will give you the title of the document and you should ask questions about this title one by one. When you ask a question, I give you the answer, and then you ask your next question. I’m only allowed to find the answer to your questions from this document, so if I cannot find the answer, I will say “I cannot find the answer, please ask your next question”.
You shouldn't ask questions that can be answered from my previous answers to your previous questions. You should sometimes ask follow-up questions from my previous answers. }
     \newline \newline
     Topic: [$t$]  \newline
        Background knowledge [$b$] \newline 
        Please start asking question about: [$h$]  \\ 
         \hline
         
    \end{tabular}
    \label{tab:prompt-temp}
\end{table*}

\subsection{Teacher simulation}
\header{Answer generation (\AGfunc).}
This component belongs to \teacherSim and is initialized with \teacherPrompt in a zero-shot manner. The \teacherPrompt includes the instruction to copy the exact spans from $s$ to answer the given question and some information about the Wikipedia page including the title $t$, background $b$, and section text $s$. We instruct \teacherSim to generate the sentence \textit{``I cannot find the answer.''} when $s$ does not contain the answer. Additionally, to prevent the generation of excessively long answers that could potentially impede readability, we implement a two-step mechanism to control the length of the generated answers:
\begin{enumerate*}[label=(\roman*)]
    \item we specify in the prompt that the selected span should not exceed a maximum of 40 tokens;
    \item we include the statement \textit{``Remember that you should select the shortest possible span from the text,''} at the end of each question, making \teacherSim itself decide on the length of the sentence within the maximum limit. 
\end{enumerate*}
 
\header{Answer validation \& regeneration (\ValidTeacher).}
Rather than solely rely on the instruction prompt \teacherPrompt for one-time answer generation, we adopt an iterative model \ValidTeacher to validate and refine the generated answers in succession to ensure they are in line with the request of \teacherPrompt. This component serves as a reminder to \AGfunc of the validation criteria and prompts \teacher to generate an answer that aligns with the given section.

We define that a \textit{valid answer} ($a_i$) should include exact copies of contiguous spans in the section text \textSection, or it should be the phrase \textit{``I cannot find the answer,''} if the question ($q_i$) cannot be answered from the text. 
Therefore, we verify an answer's validity based on two criteria:
\begin{enumerate*}[label=(\roman*),nosep,leftmargin=*]
    \item \textit{whether $a_i$ contains one or multiple exact copies of the text spans in \textSection or being ``I cannot find the answer''}; and \label{val1}
    \item \textit{whether $a_i$ is copied from the text section \textSection, rather than the background $b$}. \label{val2}
\end{enumerate*}

We follow the steps below to address the two validation criteria. First of all, to address criterion \ref{val1}, we conduct a simple text search and see if $a_i$ (or each sentence of $a_i$) is from \textSection. Notably, we notice that most of the time \acp{LLM} do not copy the texts inside the brackets and neglect the extra white spaces within the text. Therefore, we generate two normalized versions of \textSection by 
\begin{enumerate*}[label=(\roman*)]
    \item removing the extra white spaces and
    \item texts inside the brackets.
\end{enumerate*}
If $a_i$ is not found in \textSection and its normalized versions, we issue a second prompt ($p_{T}$) \textit{`Please copy the answer exactly from the given text,''} reminding \AGfunc where it has failed. To address criterion \ref{val2}, we also perform a  text search to see if $a_i$ is selected from $b$. In such a case, we issue a second prompt ($p_{T}$) \textit{``Please answer from the given section not the given background description,''} to remind \AGfunc of this criterion.
We continue these steps of validation and regeneration until the generated answer satisfies both validation criteria.

Finally, once the valid $a_i$ has been confirmed by \ValidTeacher, it is passed on to \studentSim, which utilizes $a_i$ to formulate the next question ($q_{i+1}$) in the conversation. However, there are cases where the loop continues for an excessive number of iterations. We terminate the loop in such cases, assuming that \teacherSim fails in finding the answer from \textSection, or the question is not answerable. Similar to \quac, we set the answer to such questions to ``\textit{I cannot find the answer.}'' This is necessary to prevent an infinite loop and ensure that the system remains efficient and responsive.

\subsection{Student simulation}
\header{Question generation (\QGfunc).}
To simulate the \student, we prompt the Question Generation \QGfunc component of \studentSim in a zero-shot manner.
With \studentPrompt, we instruct \studentSim to explore the given information ($h$ and $b$) by posing questions, under the assumption that it does not possess knowledge of \textSection.
As shown in Table \ref{tab:prompt-temp}, we include the topic $t$ and \background as well as the section header $h$ in \studentPrompt to ensure that \studentSim has some basic knowledge about the given topic.

\header{Question validation (\ValidStudent).}
To ensure that an LLM-generated question $q_i$ is structurally sound, we employ a validation step called \ValidStudent. This component serves the purpose of verifying and validating the syntactical correctness and coherence of the generated question. We observe that while $q_i$ is supposed to be exactly one question in our setting, sometimes the LLM tends to generate multiple questions in one go. To address this issue, we consider a question valid if it adheres to the following criteria: 
\begin{enumerate*}[label=(\roman*)]
    \item it should not exceed 25 words in length and 
    \item should not contain a newline character or enumerated items (e.g., 1, 2, 3).
\end{enumerate*}
This simple yet effective validation helps to filter lengthy and intricate questions, including those containing multiple sub-questions.

\header{Prompt selection for \student (\promptSelectionStudent).}
As the conversation progresses, there may be instances where the generated question $q_i$ remains unanswered from the given text ($s$) despite being relevant to the information need ($h$) and topic ($t$). For instance, \students tend to ask very specific follow-up questions that cannot be answered from \textSection (e.g., ``Was Newsom's mayoralty generally well-received by the citizens of San Francisco?''). To address this issue, it is crucial to continuously assess the ability of the teacher simulator to answer the generated question $q_i$ and make necessary adjustments to the \student prompt $p_S$ to enhance the quality of the question. The refined $p_{S}$ aids the generative component \QGfunc in generating questions that can be answered from the given information \textSection. For instance, if the response $a_i$ is ``I cannot find the answer,'' there is a higher chance that the subsequent question $q_{i+1}$ might be overly specific and cannot be answered directly from \textSection. To solve this issue, \promptSelectionStudent randomly selects one of the following guiding prompts as $p_{S}$ and passes it to \QGfunc. These guiding prompts include: 
\begin{enumerate*}[label=(\roman*)]
    \item Ask a general question and do not ask a too specific question; \label{st1}
    \item Ask a question starting with where, when, or who; \label{st2}
    \item Ask a question about what is interesting in this article; \label{st3}
    \item Ask a question about another aspect of the topic. \label{st4}
\end{enumerate*}
By utilizing these guiding prompts, we can effectively prevent the generation of overly specific questions and guide \studentSim by offering additional clues and information. This approach allows for more efficient exploration of the given information need ($h$) by the \studentSim, ultimately enhancing its overall understanding.

\section{Teacher Evaluation}
In this section, we describe our experimental methodology to evaluate the performance of \teacherSim from various perspectives, which addresses \textbf{RQ2} and \textbf{RQ3} from the \teacher perspective. Firstly, we describe the data source to perform \teacher evaluation. We then introduce the human evaluation process of the \teacher, with a particular focus on assessing the generated answers by comparing them against human-generated answers. 

\subsection{Experimental setup}
\header{Data for evaluating \teacherSim.}
To simulate the \teacher and ensure a fair comparison between the LLM- and human-generated answers, we maintain consistency in the conversation topic and questions across the comparison. In detail, we randomly select 50 conversations from the training set of \quac~\cite{Choi2018QuACQA}. From each conversation in the sampled data, we borrow the topic information and all associated questions. Following this, we pass the questions to our \teacherSim to generate the answers and then compare them with the original answers from \quac.

\header{Parameters.}
In our experiment, we adopt GPT-4~\cite{OpenAI2023GPT4TR}\footnote{\url{https://openai.com/gpt-4}} as our base \teacher and \student \ac{LLM} model. In our preliminary experiments, we explored using other \acp{LLM} such as GPT-3.5 and LLaMA~\cite{Touvron2023LLaMAOA} as \teacher. However, we found that GPT-4 is the only \ac{LLM} that can copy an exact segment of the text as an answer in a zero-shot manner (we later discuss it as a direction for future work in Section \ref{conclusion}). Other models failed in this task by either generating broken or free-text sentences that did not satisfy our requirements. In our model, we set the patience parameter of \ValidTeacher to a fixed value of 4, which means the \teacher validation loop breaks after a maximum number of 4 iterations.

\begin{table}[]
\small
    \centering
        \caption{Examples of cases where answers generated by \teacherSim win the original \quac answers in each aspect.}
    \begin{tabular}{p{8cm}}
    \toprule
  \centering\arraybackslash  Correctness\\ \midrule
    \textbf{1) How old was he when he went on pilgrimage?} \newline
              \textcolor{color1}{\answerQUAC:} In 1897, \newline
              \textcolor{color2}{\answerSim:} He was twenty-eight, had been married ten years, and had an infant son with another child on the way. \newline\\ \midrule
    \centering\arraybackslash   Completeness\\ \midrule        
      \textbf{2) What shows did David Frost have?} \newline
              \textcolor{color1}{\answerQUAC: }Sunday morning interview programme Breakfast  \newline
              \textcolor{color2}{\answerSim: }Sunday morning interview programme Breakfast; Through the Keyhole; Al Jazeera English. \newline\\   \midrule

\centering\arraybackslash Naturalness\\ \midrule
      \textbf{3) Did he perform in the later 30s?} \newline
              \textcolor{color1}{\answerQUAC:} Agency. Mills though continued to record Ellington  \newline
              \textcolor{color2}{\answerSim:} In 1937, Ellington returned to the Cotton Club which had relocated to the mid-town Theater District.\newline\\  \bottomrule
    \end{tabular}
    \label{tab:data-example}
\end{table}

\header{Human evaluation.} 
To evaluate the performance of the \teacher in our task, we conduct human evaluation on a professional crowd-sourcing platform Prolific.\footnote{\url{https://prolific.co}} We ask the crowd-workers to compare the answers generated by \teacherSim (i.e., \answerSim) with the answers of \quac (i.e., \answerQUAC) in terms of \textit{correctness}, \textit{completeness}, and \textit{naturalness}. 
We explain each aspect in detail:
\begin{itemize}[leftmargin=*, nosep]
    \item \textit{Correctness} aims to determine whether the selected text span accurately serves as a correct answer to the question, based on the context of the conversation.
    \item \textit{Naturalness} measures the fluency and human-likeliness of a text span. Although both \quac and \teacherSim contain a selected text span as a response, we observe that in many cases of \quac, the selected spans are unnatural and do not form complete sentences.
    \item \textit{Completeness} measures whether the provided answer is complete and comprehensive. It is important to note that an answer can be correct but incomplete. For example, if the question is about the albums of an artist, a more complete answer is the one that lists more albums, if not all.
\end{itemize}
 
Additionally, we ask the crowd-workers to indicate which system (human in \quac vs.\ \teacherSim) they would \textit{prefer} to interact with, aiming to capture the overall quality of the generated data in a conversation. Also, we ask them to provide a short statement justifying their preference. 

\header{Crowdsourcing task design.}
We design a crowdsourcing task accordingly for the assessment between two conversations.
The annotators begin by comparing the responses from both systems for each question.
We display the background information (\background) and the section text (\textSection) on the left side of the page. On the right side, we include each question along with the simulated answer (\answerSim) and the original \quac answer (\answerQUAC). For each annotation aspect, we ask the annotators to indicate which system is better by choosing from the four options, namely, ``System A,'' ``System B,'' ``Neither A nor B,'' and ``Both A and B.'' The annotators can easily locate the selected text spans by clicking on the answers. The text will be highlighted in \textSection, enabling them to compare the two spans efficiently and easily.
Note that we do not ask the annotators to evaluate the questions when the answers from \answerQUAC and \answerSim are identical. However, we still include them in the interface as they can contribute to the context of the conversation. Also, when one of the answers is ``I cannot find the answer,'' we only ask the annotators to evaluate its correctness, as other metrics cannot be evaluated for these cases.

\header{Annotation and quality check.}
We randomly sampled 50 conversations from two datasets and divided them into 10 batches, each containing five conversations for evaluation. To ensure reliable assessments, we have a minimum of three crowd-workers evaluate each conversation independently. We consider one system to have \textit{won} over the other when the majority of the crowd-workers choose it. However, we acknowledge that there may be instances where the two systems perform equally well. In such cases, no system receives the majority vote, leading to a \textit{tie}. In Table~\ref{tab:data-example}, we provide several cases when the LLM answer \answerSim wins the human answer \answerQUAC under different aspects.

To avoid any position bias in the annotations, both \quac and \teacherSim examples are randomly switched and positioned as System A and System B for each conversation.
Also, to ensure English proficiency, we made the task visible only to native English speakers. 
Additionally, before starting the annotation task, we asked the crowd-workers to complete an onboarding test, consisting of some questions about the task itself (e.g., (i) What does ``System A is correct'' mean? and (ii) Is a correct answer always natural?). We also provided around 10 sample annotations for the crowd-workers to refer to.
Upon completion, we evaluated their responses, and only if they answered at least 75\% of the onboarding questions correctly, they were allowed to start the main annotation task. This approach helps to guarantee that crowd-workers are adequately prepared and knowledgeable before undertaking the annotation tasks. 
Moreover, we manually check the consistency of preference justifications with the labels by reading their open comments. We noticed that in some cases (7\%) they do not match, so we removed them from our dataset.

\subsection{Experimental results}
In this section, we evaluate the performance of \teacherSim.

\header{Answer comparison: \quac vs.~\teacherSim.}
We report the performance on 359 questions extracted from the 50 sampled conversations. 
For 77 questions (21.4\%) \answerSim is identical to \answerQUAC.
Furthermore, for 106 of the questions (29.5\%), there is an overlap between \answerSim and \answerQUAC, indicating that one is a substring of the other.
For 176 questions (49.0\%), \answerQUAC and the \answerSim do not overlap. Notably, for 41 questions, \teacherSim returns more than one segment from the text as the answer. The statistics on comparison of \answerQUAC and \answerSim can be found in Table~\ref{Tab:teacher-res-stat}. \looseness=-1

\begin{table}%
\centering
\caption{Statistics of comparison on \answerQUAC and \answerSim under different conditions based on their answer span and type. The ``I cannot find the answer'' answers are represented by `None.' We refer to the single-span answers generated by \answerSim by `single'.}
\begin{tabular}{cp{3.5cm}cc} 
 \toprule
   \textbf{Ans.~Span} &  \textbf{Condition}  &  \textbf{Count} &  \textbf{Total} \\ 
 \midrule
\multirow{2}{*}{Overlap}  &  \answerSim is single  &   87 &\multirow{2}{*}{106 (29.5\%)} \\ \cline{2-3}
   &  \answerSim is not single  &  19  &    \\
\midrule
\multirow{6}{*}{Different}   &  \answerQUAC = None \newline AND \answerSim != None &   \multirow{2}{*}{18} &\multirow{6}{*}{176 (49.0\%)}    \\ \cline{2-3}
   &  \answerSim = None \newline AND \answerQUAC != None  &  \multirow{2}{*}{54}  &    \\   \cline{2-3}
   &  \answerSim is single  & 85   &    \\  \cline{2-3}
   &  \answerSim is not single  & 19   &    \\
\midrule
\multirow{4}{*}{Same}   &  \answerSim = None \newline AND \answerQUAC = None  &  \multirow{2}{*}{41}  & \multirow{4}{*}{77 (21.4\%)}   \\ \cline{2-3}
   &  \answerSim is single  &  33  &    \\ \cline{2-3}
   &  \answerSim is not single  &  3  &    \\
\bottomrule
\end{tabular}
\label{Tab:teacher-res-stat} 
\end{table}

\begin{table}%
\centering
\small
\caption{Results of the pairwise human evaluation of answers generated Teacher simulation  \answerSim compared to original \answerQUAC answers. Each cell reports the percentage of cases where the three human annotators agreed that either \answerQUAC or \answerSim wins. We also report the percentage of ties, where the annotators disagreed on a winner.}
\begin{tabular}{lllll} 
\toprule
    \textbf{Annot.~level}    &    \textbf{Metric}    &  \textbf{\answerQUAC}    & \textbf{\answerSim}  & \textbf{Tie} \\ 
\midrule
\multirow{3}{*}{Question} & Correctness     & 11.31\%    &  \textbf{38.6}\%   & 50.0\% \\
& Naturalness     &  7.1\%    &  \textbf{42.1}\%   & 50.7\% \\ 
& Completeness    & 5.26\%    &  \textbf{53.8}\%   & 40.8\% \\
\midrule
Conversation & Preference      &  6.12\%    &  \textbf{87.7}\%   & 6.18\%  \\
\bottomrule
   
\end{tabular}
\label{Tab:teacher-res-human} 
\end{table}

\header{Answer-level human evaluation.}
We report the result of \teacherSim human evaluation (Fleiss' $\kappa = 0.4365$) in Table~\ref{Tab:teacher-res-human}.
The result shows that \teacherSim outperforms the human \teacher of \quac in terms of all question-based metrics by a large margin. Additionally, we see that the annotators prefer the answers provided by our \teacherSim over the \answerQUAC in 87.7\% of the topics.
\teacherSim answers exhibit enhanced accuracy and naturalness due to a significant number of incomplete answer spans in \quac. This leads to grammatically incorrect sentences (e.g., ``platinum. Thank U, the',' ``Tori Amos on the 5 and a Half Weeks'').
It is also noteworthy that we allow \teacherSim to select multiple spans from the text to provide more complete answers to the questions, when necessary --- something that is missing in \quac training set, but only available in the test set~\cite{Choi2018QuACQA}. Furthermore, as in our task, we limit the LLM to answer questions from the given text, the risk of hallucination is highly decreased and is indeed verifiable, making LLMs more reliable. These findings are in line with \citet{faggioli2023perspectives}, showing the potential of LLMs in replacing crowdsourcing in annotation and simulation tasks, addressing the concerned research question \textbf{RQ3}. \looseness=-1

\header{Conversation-level human evaluation.}
We further compute \textit{Preference} as reported by the annotators, who indicate their preference for interacting with either of the two systems. 
We see in Table~\ref{Tab:teacher-res-human} that \answerSim is the winner in terms of conversation-level annotator preference, where 87.7\% of them prefer \answerSim over \answerQUAC. This indicates the promising potential of \acp{LLM} in engaging in a conversation, as long as they are sufficiently informed about the task and certain verification steps are employed.

We follow \citet{siro2022understanding} and cluster the open-ended justifications provided by the annotators into different categories to gain more insights into other aspects of quality that can be overlooked in our human evaluation. In our analysis, we find that most of the comments mention the three aspects that we include in our annotation task (i.e., correctness, naturalness, and completeness). We also find comments that can be classified into \textit{seven} new categories, namely, \textit{clarity}, \textit{coherency}, \textit{directness}, \textit{being comfortable}, \textit{trust-worthiness}, \textit{factuality}, and \textit{conciseness}. We see that many annotators found \answerSim answers more factual and found the conversations more comfortable. Interestingly, even though \answerSim answers are lengthier on average, some annotator justified their preference because of their conciseness.
\section{Simulation evaluation}
To provide a more comprehensive evaluation of \textbf{RQ2} and \textbf{RQ3}, we assess the performance of the LLM simulation in comparison to human performance. We first introduce an LLM-based simulated dataset \ourdataset. Furthermore, we report the results of state-of-the-art reading comprehension methods on the two datasets to shed light on the quality and difficulty of the simulated dataset. 

 \subsection{\ourdataset dataset}
We first introduce our dataset named \ourdataset for simulation evaluation, using the simulation framework described in Section~\ref{method}. To collect \ourdataset we used GPT-4 to implement \studentSim and \teacherSim. We randomly select 342 conversations from the training set of \quac and simulate 334 conversations using the unique topics from this sample. \ourdataset consists of 4,005 questions with an average of 1.32 answer spans per question. The statistics of \ourdataset are presented in Table~\ref{Tab:data}, alongside those of the original \quac conversations. 
\begin{table}[t]
\centering
\caption{Statistics of the collected dataset by simulating the conversation with \teacherSim and the \studentSim.}
\small
\begin{tabular}{lll} 
 \toprule
 &  \textbf{\quac} & \textbf{\ourdataset} \\ 
\midrule
\# conversations                 & \phantom{0,}342   & \phantom{0,}334   \\ 
\# questions                     &  2,498 & 4,005  \\
\# questions with answer         & 2,062  & 2,517  \\
Avg. length of the answers       & \phantom{0,0}15.33 & \phantom{0,0}28.23 \\
Avg.~\# answers per question     &  \phantom{0,00}1    & \phantom{0,00}1.32  \\

\bottomrule
   
\end{tabular}
\label{Tab:data}
\end{table}

\begin{table}[]
\footnotesize
    \centering
        \caption{The questions from \quac on ``Talland House (1882-1894)'' section of the ``Virginia Woolf'' topic with the questions generated by \studentSim.}
    \begin{tabular}{p{8cm}}
    \toprule
  \centering\arraybackslash \textcolor{color2}{\ourdataset}  \\ \midrule
 {$q_1$)} Where is Talland House located?\newline
    $q_2$) Did Virginia Woolf live in Talland House during the period of 1882-1894?\newline
    $q_3$) Who owned Talland House during this period of 1882-1894?\newline
    $q_4$) What is the architectural style of Talland House?\newline
    $q_5$) Did any notable events take place in Talland House during the period between 1882-1894?\newline
    $q_6$) What impact did living in Talland House have on Woolf's later work?
  \\\midrule
 \centering\arraybackslash  \textcolor{color1}{\quac}  \\\midrule
 {$q_1$)} What is talland house?\newline
    {$q_2$)} What happened at this house?\newline
    {$q_3$)} Did anything tragic happen?\newline
    {$q_4$)} What else happened at the house?
    \\
    \bottomrule
    \end{tabular}
    \label{tab:conv-example}
\end{table}

\vspace{-2mm}
\subsection{Student evaluation}
Due to the nature of a \student's role in a conversation, which involves asking questions and exploring a topic, it becomes challenging to define an objective metric that determines which model is ``better.'' Therefore, our emphasis lies in highlighting the distinctions between the behavior of two systems by contrasting their linguistic characteristics from various aspects.

\header{Question comparison: \quac vs.~\studentSim.}
Table~\ref{tab:conv-example} presents a sequential collection of questions in a conversation within both the \quac and \ourdataset datasets, sharing the same topic. An observation can be made that GPT-4 tends to inquire about more detailed and lengthy questions compared to humans. Additionally, it is worth noting that the human \student in \quac ceases asking questions after the fourth one, while the simulated \student in \ourdataset continues to pose additional queries. 

\begin{figure}
    \centering
    \begin{subfigure}[b]{0.45\columnwidth}
     \centering
     \includegraphics[width=\textwidth]{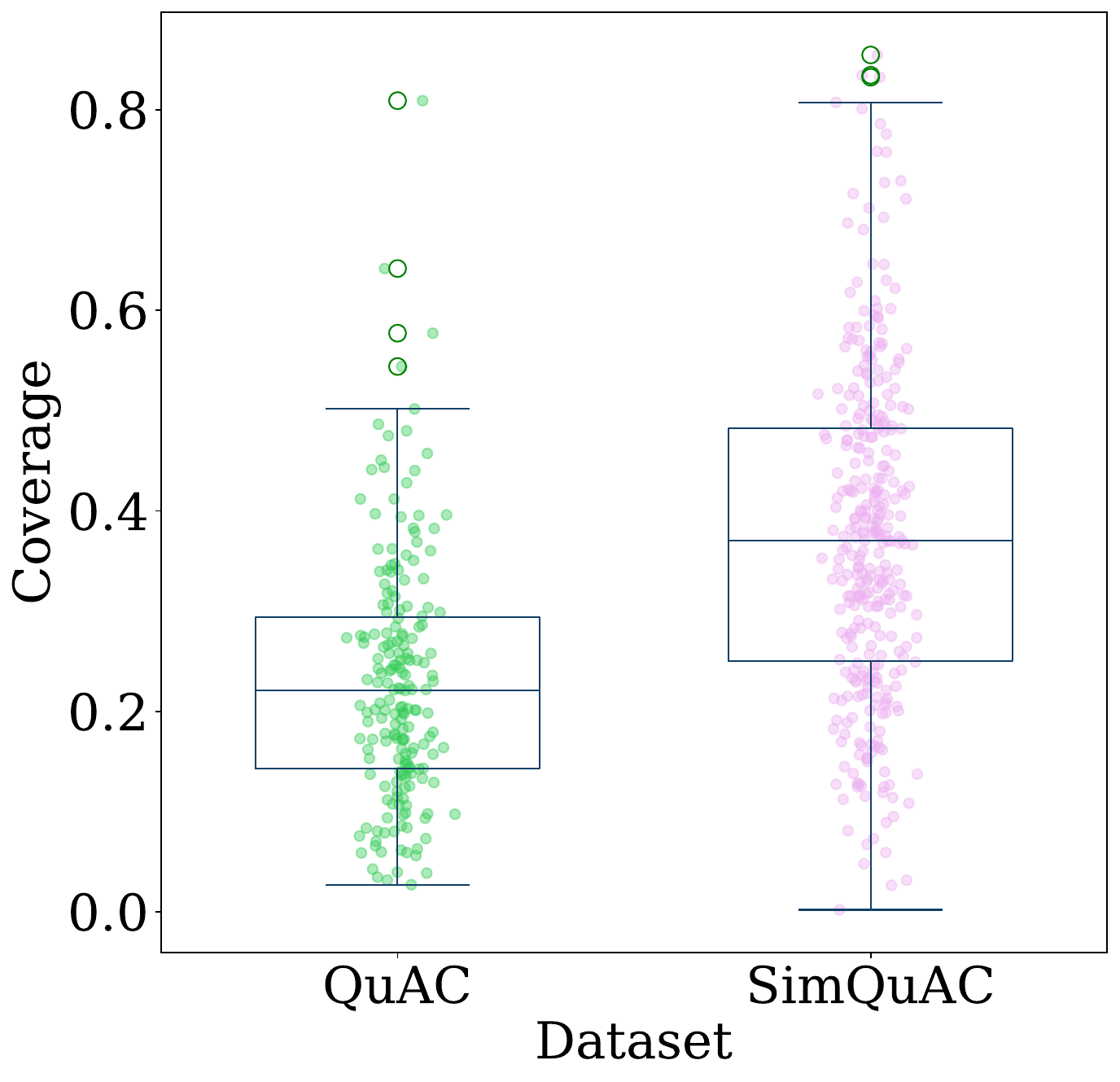}
        \vspace{-6mm}
        \caption{Topic coverage}
        \label{fig:coverage}
    \end{subfigure}%
    \begin{subfigure}[b]{0.48\columnwidth}
        \centering \includegraphics[width=\textwidth]{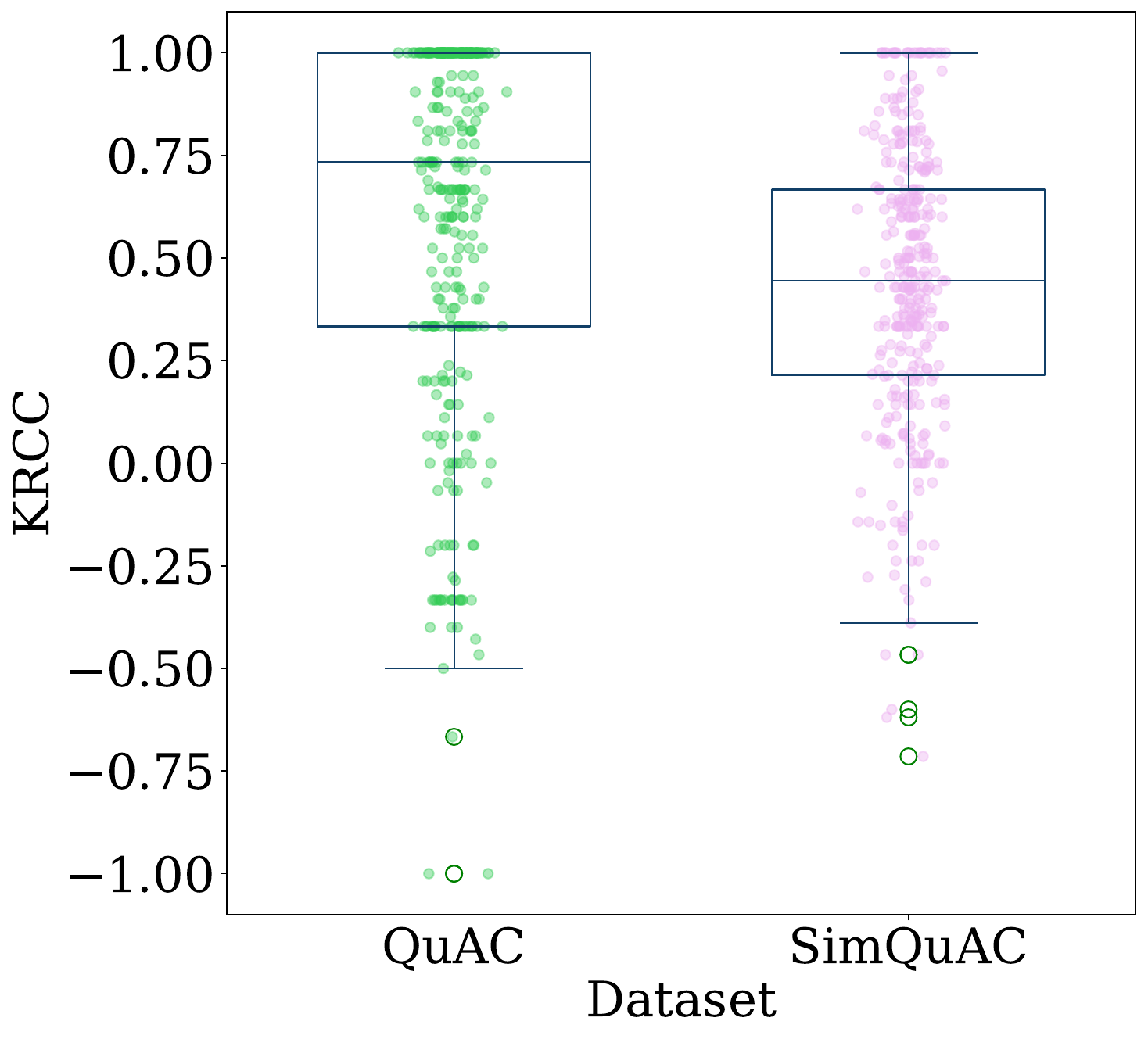}
        \vspace{-6mm}
        \caption{Conversation flow}    
        \label{fig:kendall}
    \end{subfigure}
    \caption{ Comparison between \student of \quac and \ourdataset in terms of (a) topic coverage and (b) conversation flow.}
\end{figure}

\header{Coverage.}
We assess the ability of the two \students to explore a topic by comparing how much of the \textSection is covered by the answers provided to the questions posed. 
We plot the distribution of coverage in Figure~\ref{fig:coverage}. We observe that \ourdataset questions cover a significantly (two-tailed t-test; $p$-value $ < 0.001$) larger portion of the text (mean $= 0.365$; std $= 0.163$), compared to \quac (mean $= 0.238$; std $= 0.122$), suggesting that careful prompting of \acp{LLM} can lead to a diverse and comprehensive set of questions in a conversation. 

\header{Conversation flow.}
Next, we compare the questions posed in terms of how they shape the flow of the conversation. Our objective in this experiment is to evaluate the naturalness of the conversation flow and the smoothness of topic transitions. We hypothesize that a conversation that strictly follows the sequential order of the content in \textSection is less natural. To measure this, we assign an order to the questions based on the positions of their corresponding answers in \textSection. For instance, let us consider questions A, B, and C. To determine their order, we examine the text spans of their respective answers and sort them based on the start position of each answer. In this case, say question B's answer is at the earliest position of \textSection, followed by A and C. Therefore, the question order would be \{B, A, C\}. To assess the sequential nature of the conversation flow, we compare the order of the questions in the conversation to their order in the document, considering their corresponding answers. In our example, if questions \{A, B, C\} appear in the conversation in the same order, i.e., A is followed by B and then C, the conversation flow would be considered completely linear. %

To evaluate the degree of correlation between the question order in the conversation and the corresponding answer order, we calculate the Kendall rank correlation coefficient (KRCC) metric~\cite{Abdi2007TheKR} for each conversation. KRCC measures the distance between two ranked lists, where a lower value indicates more distance between the two lists. In our case, the lower the value, the less sequential a conversation flow is.
Figure~\ref{fig:kendall} plots the distribution of the two datasets in terms of KRCC. We can see that the average value of KRCC is lower for \ourdataset than \quac, indicating that the \student of \quac poses more questions in a sequential order, compared with \studentSim. This suggests that \studentSim tends to explore the topic by jumping from one part to another part. While there is no indication as to which order is more natural, we can see that there is a clear difference in their behavior. It is noteworthy that exhibiting a more random behavior in posing questions can lead to more challenging datasets, as it prevents model learning such a biased behavior of \student.

\vspace{-2mm}
\subsection{Reading comprehension benchmarking}
To gain a deeper understanding of the distinctions between the human-generated (\quac) and the LLM-simulated data (\ourdataset), we utilize several pre-trained discriminative and generative reading comprehension baselines for evaluating the \teacher model. These models are pre-trained on the SQuAD dataset~\cite{Rajpurkar2016SQuAD1Q} and we test them on two datasets directly without further fine-tuning. 
To ensure a fair comparison with the \quac subset, we impose a limitation on \ourdataset, restricting it to a maximum of 3 questions within a conversation that do not have an answer. 
Table~\ref{tab:bert} reports the results of different models in terms of exact match (EM), precision, recall, and F1-measure when testing on the two datasets. 

The results demonstrate that, in comparison to \quac, most models exhibit superior overall performance when tested on \ourdataset. This suggests that the LLM-simulated data may provide a more favorable context for these models, leading to improved results. Furthermore, it is noteworthy that the EM score in \ourdataset is lower compared with \quac. This discrepancy can be attributed to the fact that, in \ourdataset, the answers generated by LLM tend to cover a longer span compared to the answers in \quac, posing more challenges for matching. Moreover, there are more questions with no answer in \ourdataset. The pre-trained models always output a span as an answer, instead of predicting no answer, leading to a lower EM measure. Additionally, our observations reveal the superior performance of generative methods, such as T5, compared to discriminative methods, such as BERT. This finding emphasizes the importance of utilizing generative LLMs  for this particular task. 

\begin{table}[]
    \centering
    \caption{Experimental results of reading comprehension models on \quac and \ourdataset in terms of precision (Pre.), recall (Rec.), F1-measure (F1), and exact match (EM). `-b' refers to the `-base' variant of the models, while `-l' refers to their `-large' variants. All the numbers are shown in percentages.}
    
    \begin{tabular}{p{1.7cm}p{0.4cm}p{0.5cm}p{0.4cm}p{0.3cm}cp{0.5cm}p{0.5cm}p{0.5cm}p{0.3cm}}
    \toprule
        & \multicolumn{4}{c}{\textbf{QuAC }} &&\multicolumn{4}{c}
        {\textbf{\ourdataset}} \\
    \cline{2-5} \cline{7-10}
         & Pre. & Rec. & F1 & EM & & Pre. & Rec. & F1 & EM  \\
    \midrule
   
        DistilBERT~\cite{Sanh2019DistilBERTAD} & 10.99 & 7.36 & 6.70 & 1.88 & & 15.28 & 7.59 & 7.75 & 1.36\\
       BERT-b~\cite{Devlin2019BERTPO}  & 12.87 & {10.64} & 8.31 & {2.38} & & {17.76} & 10.25 & {9.16} & 1.21    \\
       BERT-l~\cite{Devlin2019BERTPO}  & 24.93 & {18.16} & 16.33 & {4.17} & & {29.74} & 16.84 & {16.75} & 2.26     \\
     \midrule
        T5-b~\cite{Raffel2019ExploringTL}  & 26.69 & {21.58} & {18.93} & {6.28} & & {31.63} & 18.02 & 18.13 & 2.52    \\
        T5-l~\cite{Raffel2019ExploringTL}  & 29.70 & \textbf{23.45} & \textbf{21.26} & \textbf{7.27} & & \textbf{35.63} & 20.90 & 21.24 & 3.01     \\
    \bottomrule
    \end{tabular}
    \label{tab:bert}
\end{table}

\vspace{-2mm}
\section{Related Work}
\subsection{Conversational question answering}
\Ac{CQA} requires the ability to correctly interpret a question in the context of previous conversation turns~\cite{Zaib2021ConversationalQA}. Under this context, modern \ac{CQA} systems can be divided into two types: sequential knowledge-based question-answering (KB-QA) agents~\cite{Iyyer2017SearchbasedNS,Saha2018ComplexSQ,Christmann2019LookBY} and conversational machine reading comprehension (CMRC) systems~\cite{Reddy2018CoQAAC,Huang2018FlowQAGF,Yeh2019FlowDeltaMF,Li2019AUM,Qu2019BERTWH}. In sequential KB-QA systems, agents need to search the database for the appropriate information to generate the answer. In this paper, we focus on the CMRC setting, where the conversation revolves around a given article and the answers are typically a span in the given resource. To this end, several datasets such as CoQA~\cite{Reddy2018CoQAAC}, FlowQA~\cite{Huang2018FlowQAGF}, have been proposed. 
Among all the CMRC datasets, \quac contains over 14K crowdsourced QA dialogs~\cite{Choi2018QuACQA}. This dataset allows for a \student posing a sequence of free-form questions to learn as much as possible about a hidden Wikipedia article and a \teacher is hired to find the answer to each question in the text. Following this line, extensive tasks such as reading comprehension~\cite{Huang2018FlowQAGF,Zhao2021RoRRF,qian2022capturing,Qu2019BERTWH}, answer ranking~\cite{Qu2020OpenRetrievalCQ}, question generation~\cite{Kumar2023ImprovingRC,Xu2022FantasticQA} are adopted to measure the performance on both \student and \teacher levels.

\vspace{-2mm}
\subsection{User simulation}
While user simulators have been studied in the \ac{IR} community extensively~\cite{balog2023user,mostafa2003simulation,carterette2011simulating}, including applications such as simulating user satisfaction for the evaluation of task-oriented dialogue systems~\cite{sun2021simulating} and recommender systems~\cite{zhang2020evaluating,afzali2023usersimcrs}, they are often limited to reacting to a system's action.
The emergence of LLMs provides the opportunity to improve user simulation, making it more realistic. LLMs are ideal for human simulations due to their remarkable ability to process text in the natural language format. They are also able to generate coherent and contextually appropriate language that is very similar to how humans communicate~\cite{Park2023GenerativeAI,Park2022SocialSC,Zhao2023ASO}. One such application is using LLMs as evaluators to mimic human evaluation, which has proven to be highly effective in various contexts~\cite{Pegoraro2023ToCO,Guo2023HowCI,Zhu2023CanCR,Qin2023IsCA,Aiyappa2023CanWT,Schick2023ToolformerLM}. For instance, ~\citet{Guo2023HowCI} compare ChatGPT with human experts by collecting tens of thousands of comparison responses from both sources. \citet{Tan2023EvaluationOC} assess the performance of ChatGPT as a KB-QA system using its own knowledge. Another common application of LLMs in simulation is leveraging them as an annotator-free tool for data augmentation~\cite{Kuzman2023ChatGPTBO,Huang2023IsCB,Askari2023GeneratingSD,askari2023test,Hu2023UnlockingTP,Tang2023DoesSD,Whitehouse2023LLMpoweredDA}.
\citet{sekulic2022evaluating} employ GPT-2 and propose an evaluation framework based on mixed-initiative conversations. \citet{owoicho2023exploiting} take it one step further and utilize GPT-3.5 to simulate a user that can also provide feedback on the relevance of a returned document in a conversational search setting. 
In the context of document re-ranking, ~\citet{Askari2023GeneratingSD} prompt LLMs to generate synthetic training data for cross-encoder re-rankers. Most recently, ~\citet{Hu2023UnlockingTP} adopt LLMs as user simulators in a task-oriented dialogue system. To the best of our knowledge, our work is the first to utilize LLMs as annotator-free \teacher--\student simulators in a \ac{CQA} system, where the \student takes a proactive role in exploring a topic.

\section{Conclusions and future work}
\label{conclusion}
We explore simulating human-to-human conversations using zero-shot prompting of LLMs in a question-answering setting. Our framework involves two GPT-4s interacting on a topic: one as the student generating questions based on background knowledge, and the other as the teacher seeking answers within a text on the given topic. To assess the system, we initially evaluate the \teacher's performance through both automated methods and human assessment. Subsequently, we compare conversations generated by the LLM and those by humans for the student-level evaluation. In summary, our investigation highlights the potential of LLMs in facilitating interactive and informative retrieval experiences.

Despite the superiority of our model, several limitations persist, which point to avenues for future research. Firstly, according to our findings, only GPT-4 consistently follows instructions to generate reasonable conversations, constraining the overall effectiveness of the pipeline. Besides, language models can exhibit various biases, when used for such simulation. It therefore becomes essential to further develop methods to mitigate these biases. Moreover, although we have devised prompting strategies to mimic human interaction, the manual construction of instructions can be time-consuming. Future work should explore more advanced and efficient automatic prompting strategies to enhance the system.
\section{Ethical Considerations}
Our work revolves around utilizing LLMs to simulate users. Given the recent emergence of LLMs and the vast interest in using them for various research directions, we believe that pursuing such a direction is necessary as it unveils the potential of LLMs, while at the same time exhibiting their potential ethical considerations. 
Below we list some of these concerns that need to be considered and addressed in this research area:
\begin{itemize}[nosep,leftmargin=*]
    \item Bias and discrimination: LLMs are biased towards their training data. The simulated data could in turn carry the same biases and further propagate stereotypes and discrimination.
    \item Misrepresentation: using an LLM to simulate users would introduce certain biases on the type of users being represented. The biases that exist in the data that the LLM is trained on would be reflected in the simulated user.
    \item Transparency and accountability: the decision-making process within LLMs can be opaque, making it challenging to understand how or why a particular simulated conversation is generated. This lack of transparency can lead to ethical challenges, particularly in contexts where clear justification for a decision is required.
    \item Environmental impact: the training and operation of LLMs consume significant computational resources, contributing to energy consumption and potentially having a negative environmental impact.
\end{itemize}
While simulating users using LLMs has various advantages, it must be approached with careful consideration of the potential ethical implications.

\bibliographystyle{ACM-Reference-Format}
\balance
\bibliography{references}

\clearpage
\end{document}